\RequirePackage{fix-cm}
\documentclass[twocolumn]{svjour3}          
\smartqed  
\usepackage{graphicx}
\usepackage{amsmath}

\DeclareMathOperator*{\argmax}{argmax}
\DeclareMathOperator*{\argmin}{argmin}
\usepackage[english]{babel}
\usepackage[utf8]{inputenc}
\usepackage{algorithm}
\usepackage[noend]{algpseudocode}
\usepackage{tikz}
\usepackage{caption}
\usetikzlibrary{shapes.geometric, arrows}
\usepackage{hyperref}
\usepackage{enumitem}
\usepackage{caption}
\usepackage{subcaption}
\newlist{steps}{enumerate}{1}
\setlist[steps, 1]{label = Step \arabic*:}
\usepackage{adjustbox}
\usepackage{geometry}
\journalname{}
\begin{document}

\title{A Distributed Approximate Nearest Neighbor Method for  Real-Time Face Recognition
}


\author{Aysan Aghazadeh         \and
        Maryam Amirmazlaghani\textsuperscript{*} 
}


\institute{Aysan Aghazadeh \at
              No. 424, Hafez Ave., Tehran, Iran  \\
              \email{$aisan\_aghazade@aut.ac.ir$}           
           \and
           Maryam Amirmazlaghani \at
              No. 424, Hafez Ave., Tehran, Iran  \\
              \email{mazlaghani@aut.ac.ir} 
}

\date{Received: date / Accepted: date}

\maketitle

\begin{abstract}
Nowadays\color{black}, \color{black}face recognition and more generally image recognition have many applications in the modern world and are widely used in our daily tasks. \color{black}This paper aims to \color{black}propose a distributed approximate nearest neighbor (ANN) method for 
real-time face recognition \color{black}using a \color{black}big dataset that involves a lot of classes.
The proposed approach is based on using a clustering method to separate the dataset into different clusters and \color{black}on \color{black}specifying  the importance of each cluster by defining cluster weights. \color{black} To this end, r\color{black}eference instances  are selected from each cluster based on the cluster weights using a maximum likelihood approach. This process leads to a more informed selection of instances, so \color{black}it \color{black}enhances the performance of the algorithm. Experimental results confirm the efficiency of the proposed method and its out-performance in terms of accuracy and \color{black} the \color{black}processing time.

\keywords{Approximate Nearest Neighbor \and Face Recognition \and Clustering \and Maximum Likelihood}
\end{abstract}

\section{Introduction}
\label{intro}
Face \color{black}recognition and image recognition \color{black}are two important fields in computer science and mainly \color{black}in \color{black}machine learning. Their relevance in the modern world requirements is \color{black}demonstrated \color{black}as a case of human machine interaction\color{black}s\color{black} \cite{Perf_PCA}\color{black}. In addition, t\color{black}hey have many applications \color{black}in fields, such as access control and security, health-care, advertising, criminal identification, \color{black} etc. \cite{face-recognition-application}. In all aforementioned applications, there are several classes, some reference images for each class\color{black}, \color{black}and one or more input images. The purpose of this type of problem is \color{black}to define \color{black}the \color{black}the class most fitting \color{black}to the input image based on the \color{black}information acquired \color{black}from reference images.\\ 

\color{black}
In all image recognition tasks, the first step is to extract features. In eager algorithms, training is the following step. Next, the features of the input image must be extracted. Finally, the most probable class is identified for the input image.\color{black}\\
\color{black}Feature Extraction, in all machine learning problems, is one of the most important and challenging steps \cite{ML-ANN}. \color{black}In face recognition and generally \color{black}in \color{black}image recognition problems, this process is more challenging due to the effect of lighting and complex background variability \cite{Novel_Haar} \color{black} as well as \color{black}variability of objects presented in images \cite{ML-ANN-1}.\\ There are many \color{black}feature extraction methods \color{black}such as \color{black}SIFT\footnote{\color{black}Scale Invariant Feature Transform} \cite{SIFT}, which was applied in \cite{sift-application}, SURF\footnote{\color{black}Speeded Up Robust Features} \cite{SURF} as applied in \cite{surf-application}, and HOG\footnote{\color{black}Histogram of Oriented Gradient} \cite{HOG} as applied in \cite{hog-application}. \color{black} Based on the application, we \color{black}could decide which one would be used.\color{black}\\
After extracting the feature\color{black}s \color{black} of all images, it \color{black}would be \color{black}time to use machine learning algorithms \cite{ML-ANN} to classify the input image. Face recognition is \color{black}an \color{black}important and challenging task\color{black}, especially when \color{black}working with big datasets \cite{DML}. \color{black}Depending on \color{black}the type of \color{black}the \color{black}algorithm used in this task, there \color{black}could \color{black}be a need \color{black}for training \color{black} reference images. For example, \color{black}neural networks are among the most useful methods in face recognition problems that include training phase. There are other types of algorithm that do not include this phase, which are called lazy algorithms\color{black}. One of the examples \color{black}of \color{black}this type of algorithm is the nearest neighbour algorithm. In \color{black}this \color{black}type of algorithm, due to the elimination of the training phase, \color{black}the \color{black}test phase \color{black}becomes \color{black}more important and time-consuming. \color{black}Thus, \color{black}one of the challenges in this \color{black}type \color{black}of algorithm is \color{black}to reduce \color{black}the processing time.\\
Finally, the output \color{black}is \color{black}a class label assigned to the input image.\\
As mentioned before\color{black}, \color{black}we \color{black}could \color{black}use many algorithms in the face and object recognition process, \color{black}with some of them being more frequently used, such as the Deep Neural Network \cite{DNN} or the Support Vector Machine \cite{ML-ANN}. \color{black}Although both of these algorithms are efficient \color{black}when the size of \color{black}the \color{black}training set is large, they are not accurate enough in problems with few reference images per class \cite{ML-ANN-12}. \color{black}Problems with the small size of training set are\color{black} important in face recognition, because sometimes, \color{black}there is \color{black}only one reference image for each person \cite{ML-ANN}. In these types of problems, \color{black} NN\footnote{Nearest Neighbor} algorithms could be used\color{black}; however, \color{black}if the number of classes is large\color{black}, \color{black}the total number of reference images will be large \color{black}as well\color{black}. \color{black}Thus, considering all reference images when choosing the nearest one would be time-consuming\color{black}. In these cases\color{black}, \color{black}the algorithm \color{black}would \color{black}be \color{black}slow \cite{ML-ANN} \cite{DEM} so \color{black} it would not be an \color{black}appropriate choice for real-time applications\color{black}, \color{black}such as face recognition. To overcome this drawback, ANN\footnote{Approximate Nearest Neighbor} algorithms have been proposed.\color{black} There are many \color{black} ANN  algorithms such as DEM\footnote{Directed Enumeration Method} \cite{DEM} and \color{black} ML-ANN\footnote{Maximum Likelihood Approximate Nearest Neighbor} \cite{ML-ANN} \color{black} that will be considered in section \ref{related-researches}\color{black}. \color{black} Both of them are \color{black}faster than the NN algorithm, but yet there \color{black}could \color{black}be a time problem for large data, and that is because of the procedure of choosing the reference images to be considered. \color{black}However\color{black}, in both approaches, choosing the instances in each iteration is based on the information acquired from \color{black}the images considered \color{black}in previous iterations. \color{black}Thus\color{black}, information obtained from the considered images is highly dependent on the first instance chosen randomly in both algorithms. \color{black}In this paper, to overcome the weaknesses of \color{black}the \color{black}ML-ANN approach, we propose a distributed version of ML-ANN. The proposed approach is based on clustering reference images\color{black}, which is \color{black}faster than ML-ANN. \\
\color{black}This paper is organized as follows: In section \ref{related-researches} we will review the related researches and explain the ML-ANN algorithm in detail; the proposed method will be described in section \ref{sec:2}. The experimental results will be presented and compared in section \ref{sec:3}; in the end, concluding remarks will be presented in section \ref{sec:4}.\color{black}
\section{Related Researches}
\label{related-researches}
In this section, we \color{black}review \color{black}the various types of recognition methods and mainly \color{black} NN \color{black} algorithms\color{black},as well as their advantages and disadvantages in different cases.\\\color{black}
\color{black}As mentioned in the previous section, for datasets having few images per each class, due to the lack of accuracy, the deep learning algorithm \cite{deep-learning} or SVM \cite{SVM} could not be used. In these types of problems, \color{black}the NN algorithms could be used.\color{black} \\
\color{black}NN algorithms are classified into\color{black}: exact \cite{ML-ANN-28}, and approximate \cite{DEM-16} \cite{ML-ANN}. \color{black}The exact form of NN algorithm search \color{black}for the exact nearest neighbor without considering all the images \cite{ML-ANN}. One of the most \color{black}commonly used methods among the exact form of NN algorithms is using kd-tree \cite{ML-ANN-28}\color{black}. \color{black}By using this data structure, this method \color{black}could \color{black}find the nearest neighbor \color{black} just like \color{black} using \color{black} a binary search tree, but with additional backtracking \cite{ML-ANN}. \color{black}Some branches are eliminated to speed up the algorithm. This data structure could be used \color{black} when the data are low-dimensional\color{black}; however \color{black}when \color{black}the \color{black}data are high-dimensional, backtracking will not eliminate the branches, and \color{black}the use of \color{black}this structure will not be efficient.\\
\color{black}In situations like the ones mentioned above \color{black}where \color{black}the exact NN methods are not efficient, \color{black}an \color{black}approximate form \color{black}could \color{black}be used \color{black}. In this type of algorithms instead of searching for the exact nearest neighbor, \color{black}the method tries \color{black} to find the neighbor \color{black}within an \color{black}acceptable distance \color{black}from \color{black}the input image.\\
There are many types of \color{black} ANN \color{black} algorithms, such as \color{black}the \color{black}clustering \color{black}of \color{black}reference images and using a hierarchical tree \cite{DEM-16}\and\cite{DEM}. The nodes in each layer are the centroids of the lower layers. In each layer\color{black}, \color{black} the nearest node is selected\color{black}, \color{black}and so in the lower layer\color{black}, \color{black} images in the approximate nearest cluster are considered. This algorithm is so fast but not \color{black}that \color{black}much accurate \cite{DEM}.\\
There is another type of approximate algorithm \color{black}that \color{black}defines a threshold and searches for the neighbor with the distance lower than the threshold. In this form of algorithm, there are two important factors which should be considered, \color{black}with the first one being the way \color{black}the threshold is defined\color{black}. \color{black}To define threshold, \color{black}type 2 error is usually used\color{black}.  \footnote{One could see the procedure of defining the threshold in \cite{DEM} in detail.}. The other factor is the order \color{black}of checking \color{black}the reference images. \color{black}It is of high importance the way the reference images are selected, with different selection procedures resulting in different ANN algorithms.\color{black}\\
One of the algorithms of this type is DEM\footnote{Directed Enumeration Method} \cite{DEM}. \color{black}DEM, a matrix of distances between all reference images is created \color{black}before starting the recognition process and \color{black}used \color{black}it in the whole process. In this method, the first instance is selected randomly, and the others are chosen based on their distances from the nearest considered instance \color{black} selected\color{black}. In each step, all the reference images which have a lower distance \color{black}from \color{black}the nearest considered image than the distance between the nearest considered image and the input \color{black}one\color{black}. This process is terminated \color{black}by determining \color{black}whether there is an instance \color{black}with a \color{black}lower distance with the input image than the threshold or \color{black}by determining \color{black}the number of iterations crossing the previously defined maximum number of iterations.\\ The problem of this algorithm is that \color{black}the images are selected \color{black}only based on their distances from the nearest considered image, \color{black}without using \color{black}the information acquired from other considered instances. \color{black}To solve \color{black}this problem, the likelihood \color{black}of \color{black}each reference image \color{black}could be calculated \color{black}based on the entire information \color{black}gathered \color{black}from the considered images. This procedure is used in \color{black} ML-ANN \color{black} \cite{ML-ANN}, which we \color{black}will explain in detail in the \ref{sub-sec:1}.\color{black}

\subsection{ML-ANN Algorithm}
\label{sub-sec:1}
Previously,  \color{black}different NN  and ANN algorithms were considered. Here, \color{black}as the method of this study \color{black}is based on the ML-ANN algorithm, we describe it in more detail \color{black}in this section. \color{black}\\
The goal of the \color{black}image recognition problem is \color{black}to find \color{black}the best fitting class for the observed input image. \color{black}It is assumed for all images that both reference images and the input image have the same width of  U and height of V\cite{ML-ANN}. The \color{black} ML-ANN \color{black}algorithm is based on using HOG\footnote{Histogram of Oriented Gradients} \cite{HOG} features.\\
It should be \color{black}noted \color{black}that for an image sized U*V, we have N HOG features where N is the number of bins in the histogram for these features.
After extracting the features, the learning algorithm is used. In \color{black} ANN \color{black} algorithms such as \cite{DEM} and \cite{ML-ANN}, \color{black}firstly the \color{black}threshold ($\rho_0$) for the distance between reference images and the input image is determined. \color{black}Next\color{black}, instead of searching for the nearest reference image, the algorithm searches for the acceptable reference image. A reference image \color{black}whose \color{black}distance \color{black}from \color{black}the input image is smaller than the threshold could be acceptable. Such \color{black} ANN \color{black} algorithms are terminated when an acceptable reference image is found to decide the class of the input image is the same as that of the acceptable reference image.that of the acceptable reference image. \\
\color{black}Thus, \color{black}the termination conditions for these algorithms could be formulated as follows \cite{ML-ANN}:
\begin{equation}\tag{1}
    \label{eq:1}
    W_\nu: \rho(X, X_r) < \rho_0
\end{equation}
where X and $X_r$ are the input and the selected reference images, respectively, and $\rho$(.,.) denotes the distance function. 
\color{black}In \cite{ML-ANN}, the chi-square distance is used. The chi-square distance is formulated as \color{black}follows\color{black}: 
\begin{equation}\tag{2}
    \label{eq:2}
    \rho_{chi-square}(X, X_r) = \sum_{i=1}^N \frac{(X_i - {X_r}_i)^2}{{X_i+{X_r}_i}}
\end{equation}

where $X_i$ is the \color{black}$i^{th}$ \color{black}HOG feature of the input image, and ${X_r}_i$ is the  \color{black}$i^{th}$ \color{black}HOG feature of the reference image. \\
The first reference image to \color{black}be compared \color{black}with the input image, is chosen randomly. If this image \color{black}fulfills \color{black}criterion \ref{eq:1}, the process will be finished; otherwise,\color{black} another reference image should be selected to be compared \color{black}with the input image, so \color{black}a queue of chosen reference images is \color{black}to be formed\color{black}.\\
The order of images in this queue is of great importance. \color{black}The method of choosing \color{black}the next reference image is the \color{black}central theme \color{black}of the \color{black} ML-ANN \color{black} method \cite{ML-ANN}. As \color{black}its name suggests\color{black}, it chooses the most probable reference image based on the reference images checked before.\\
\color{black}Let's assume \color{black}one has \color{black}checked k reference images and now \color{black}wants \color{black}to choose $X_{k+1}$ \cite{ML-ANN}. Based on the Bayesian theorem and independence of the reference images from different classes, the maximal likelihood choice is obtained using the following formula: \cite{ML-ANN}:
\begin{equation}\tag{3}
    \label{eq:3}
    r_{k+1}=\argmax_{\nu \in \{1, ..., R\}-\{r_1, ..., r_k\}} (p_\nu . \prod_{i=1}^k f(\rho (X, X_{r_i})|W_\nu)
\end{equation}

where $x_{r_1}, ..., x_{r_k}$ are the first to $k^{th}$ considered reference images, $p_\nu$ is the probability of the $\nu^{th}$ class to be the best fitting class, and $f(\rho (X, X_r)|W_\nu)$ is the conditional density of the distance $\rho (X, X_r)$ if the hypothesis $W_\nu$ is true (the class label of observation X is $\nu$)\cite{ML-ANN}.
Since \color{black}the chi-square distance is used and the output of the HOG method is a $U \times V \times N$ matrix where N is the number of bins\color{black}, the output of the distance is a chi-square distribution with N-1 degrees of freedom \cite{ML-ANN}. After approximating the chi-square distribution with a normal distribution, $f(\rho(X, X_{r_i})|W_\nu)$ \color{black}could \color{black}be computed as\footnote{One can see the process of the approximating the distribution and computing the $f(\rho(X, X_{r_i})|W_\nu)$ in details in paper \cite{ML-ANN}}:
$$f(\rho(X, X_{r_i})|W_\nu)= \frac{UV}{\sqrt{2\pi.(4UV.\rho_{\nu, {r_i}}+2(N-1))}}+ $$
\begin{equation}\tag{4}
    \label{eq:4}
       exp[-\frac{UV(\rho(X, X_{r_i}) - \rho_{\nu, {r_i}} - \frac{N-1}{UV}) ^ 2}{8\rho_{\nu, {r_i}} + \frac{4(N-1)}{UV}}]
\end{equation}

\color{black}Thus, \ref{eq:3} could be written as follows:\color{black}

\begin{equation}\tag{5}
    \label{eq:5}
    r_{k+1} = \argmin_{\mu \in \{1, ..., R\}-\{r_1, ..., r_k\}} (\sum_{i=1}^k \varphi_\mu(r_i) - \ln p_\mu)
\end{equation}
where
\begin{equation}\tag{6}
    \label{eq:6}
    \varphi _\mu(r_i) \approx \frac{(\rho(X, X_{r_i}) - \rho_{\mu, r_i})^2}{\rho_{\mu, r_i}}
\end{equation}

In \cite{ML-ANN}, it is assumed that the matrix of the distances between all reference images is computed before starting the recognition\color{black}; thus, \color{black}there is no distance-computation in selecting the instance process, and the recognition process will be real-time. This approach encounters a problem when the input image \color{black}belongs to no classes\color{black}. \color{black}To avoid this problem, the maximum number of iterations is defined\color{black}, and \color{black}if \color{black}there is no reference \color{black}images fulfilling \color{black}the criterion \ref{eq:1}, the process will be terminated.
\section{Proposed Approach}
\label{sec:2}
Although \color{black}the ML-ANN \color{black} \cite{ML-ANN} algorithm is \color{black}an \color{black}improved form of the DEM\cite{DEM} method and works faster, \color{black}a time problem could face big data.\\
\color{black}
In \color{black}the ML-ANN \color{black} method, a reference image is selected and the termination condition is examined for it \color{black}each time\color{black}. If this condition is not \color{black}fulfilled\color{black}, the next reference image will be selected. \color{black}Thus, \color{black}working on more than one image \color{black}at the same time \color{black}is not possible in this method. \color{black}Hence we aim to \color{black}propose a method that can work on more than one reference image simultaneously; \color{black}besides\color{black}, it \color{black}is well capable of\color{black} parallelism and \color{black}could \color{black}be more useful in large scale datasets.\\
\color{black}
Selecting reference images, in \color{black}the ML-ANN \color{black} method is based on likelihoods; however, selecting reference images \color{black}should be done \color{black} more consciously\color{black}; accordingly, \color{black} this paper \color{black}aims to \color{black}propose a faster and more conscious method.\\
As mentioned in the previous section, calculating the likelihood of each reference image is based on the information acquired from previous selected images in the queue of reference images. Hence selecting the reference images depends on the first image chosen randomly.\\
In \color{black}the present \color{black}method, the \color{black} ML-ANN \color{black} algorithm \color{black}is combined \color{black}with a clustering algorithm to select the next instance more accurately. In the following \color{black}section\color{black}, we will explain the \color{black}method \color{black}in detail.\\
Similar to ML-ANN, before starting the recognition process, the matrix of distances between all reference images \color{black}is computed, which \color{black}makes the algorithms faster. In the proposed method, \color{black}apart from \color{black}creating this matrix, we cluster all the reference images as follows:
\begin{equation}\tag{7}
    \label{eq:7}
    \forall X_r,\ r\in\{1,...,R\};\ X_r \in C_i,\ i\in\{1,...,K\}
\end{equation}
Where $C_i$ denotes the \color{black}$i^{th}$ \color{black}cluster, and K is the number of clusters. 
K \color{black}could be chosen \color{black}based on the number of classes. However, K should be much less than the number of classes, i.e. each cluster consists of some classes. We use the K-means algorithm in this paper.\\
\color{black}The recognition \color{black}process starts with creating a queue for each cluster, \color{black} which is the reason \color{black}why the proposed algorithm could be run in parallel. Unlike the \color{black} ML-ANN \color{black} algorithm where the first instance is chosen randomly, in this algorithm, we use the centroid of each cluster as the first instance added to each queue to \color{black}distribute them better \color{black}on the dataset: 
\begin{equation}\tag{8}
    \label{eq:8}
    queue_{i} = \{centroid_i\}; for\ i\in \{1,...,K\}
\end{equation}
After adding the centroids to the corresponding queue, the distance between the instances and the input image is calculated, and if one of them is less than the threshold, the algorithm will be terminated\color{black}; otherwise\color{black}, we must select the next reference images from each cluster and add them to the corresponding queue. \\In this step, to make the method more efficient, an approach \color{black}is proposed \color{black}to add different numbers of reference images to different queues. \color{black}To better clarify the reason, let's \color{black}assume that the distance between the centroid of \color{black}the $i^{th}$ \color{black}cluster and the input image is small (but not smaller than the threshold), and that the distance between the centroid of \color{black}the $j^{th}$ \color{black}cluster and the input image is large. \color{black}One could \color{black}conclude it is more likely  that the input image belongs to \color{black}the $i^{th}$ \color{black}cluster and the classes inside this cluster compared to the \color{black}$j^{th}$ \color{black}cluster. \color{black}Thus\color{black}, to consider this important issue in \color{black}the \color{black}proposed method, we \color{black}assign \color{black}a weight \color{black}to \color{black}each cluster, \color{black}which \color{black}demonstrates the importance of that cluster. It \color{black}is worth noting \color{black}that the weight of each cluster changes when new instances are added to the cluster queue. In other words, the weight of each cluster, in each step, is computed based on the last version of its queue. \color{black}In addition, \color{black}the number of the instances selected from each cluster is based on the weight assigned to it. \color{black}Hence, \color{black}before choosing the next instances, the weights assigned to each cluster must be defined. \color{black}To calculate \color{black}the weights, there are many procedures such as, \color{black}the use of \color{black}the average distance between the selected instances in each queue and the input image, the maximum or minimum distance, the average likelihood of each queue, etc., to be used. For the rest of the paper, we use average distances to calculate the weights. At first, we calculate the average of the distances between the instances in each queue and the input image as follows:
\begin{equation}\tag{9}
    \label{eq:9}
    \forall j \in \{1, ..., K\}; avg(distance)_j = \frac{\sum_{i=0}^{k_j}{\rho(X, X_{j_i})}}{k_j}
\end{equation}
Where K is the number of clusters, $avg(distance)_j$ is the average of distances between instances in the \color{black}$j^{th}$ \color{black}queue and the input image, and $k_j$is the number of images in the \color{black}$k^{th}$ queue.\\
\color{black}Next, \color{black}we select the maximum value between the average distances of each queue as formulated below:
\begin{equation}\tag{10}
    \label{eq:10}
    max\_average\_dist\ =\ \max_{j \in \{1, ..., K\}}{avg(distance)_j}
\end{equation}
\color{black}Accordingly\color{black}, we divide the maximum value by the calculated average distance ($avg(distance)_j$) for each queue and use the nearest integer number above the obtained value for each queue as the weight assigned to that queue. This process is formulated as follows:
\begin{equation}\tag{11}
    \label{eq:11}
    \forall i \in \{1,...,K\}; \  weight_i \ = \ \bigg \lceil\frac{max\_average\_dist}{avg(distances)_i}\bigg\rceil
\end{equation}
\normalsize
The weight assigned to the queue with the maximum average distance equals one, and the others get their value according to their average distance. \color{black}The smaller the average distance is, the bigger the weight will be. \color{black}Having calculated the weights assigned to each queue, we must select the instances for the next iteration. \color{black}To select \color{black}the instances from each cluster in the next iteration, we use the same procedure as the \color{black} ML-ANN \color{black} method with little changes. The instances are chosen using the criterion \ref{eq:12}. \color{black}In addition, \color{black}according to formula \ref{eq:12}, to select the most probable instances for each cluster queue, all previously compared instances in all cluster queues \color{black}are used\color{black}. 
$$\forall j \in \{1,...,NC\};\ \ \ r_{(k+1)_{j}}\ =$$
\begin{equation}\tag{12}
    \label{eq:12}
     \argmin_{\mu \in \{1_j,...,R_{j}\}-\{r_{1_j},...,r_{k_{j}}\}}\bigg(\sum_{l=1}^{NC}\sum_{i=1}^{k_{l}}\varphi_\mu(r_i)-\ln p_\mu\bigg)
\end{equation}
\normalsize\\
Where $r_{(k+1)_j}$ is the index of the next selected instance from the \color{black}$j^{th}$ \color{black}cluster, $k_l$ is the number of selected instances from the \color{black}$l^{th}$ \color{black}cluster in previous iterations, and $\varphi_\mu(r_i)$ is the value calculated by the criterion \ref{eq:6}.
This process \color{black}will be terminated\color{black}, if there is an instance fulfilling the criterion \ref{eq:1}, or if the number of iterations exceeds the maximum iterations \cite{ML-ANN}. If there are more than one reference image \color{black}fulfilling \color{black}the criterion \ref{eq:1} or the number of iterations exceeds the maximum iterations, we will choose the nearest image among the considered images.
This process helps us choose reference images more consciously by \color{black}selecting \color{black}instances from different clusters.\\

\begin{figure}[ht]
    \centering
    \begin{center}
        \includegraphics[scale = 0.23]{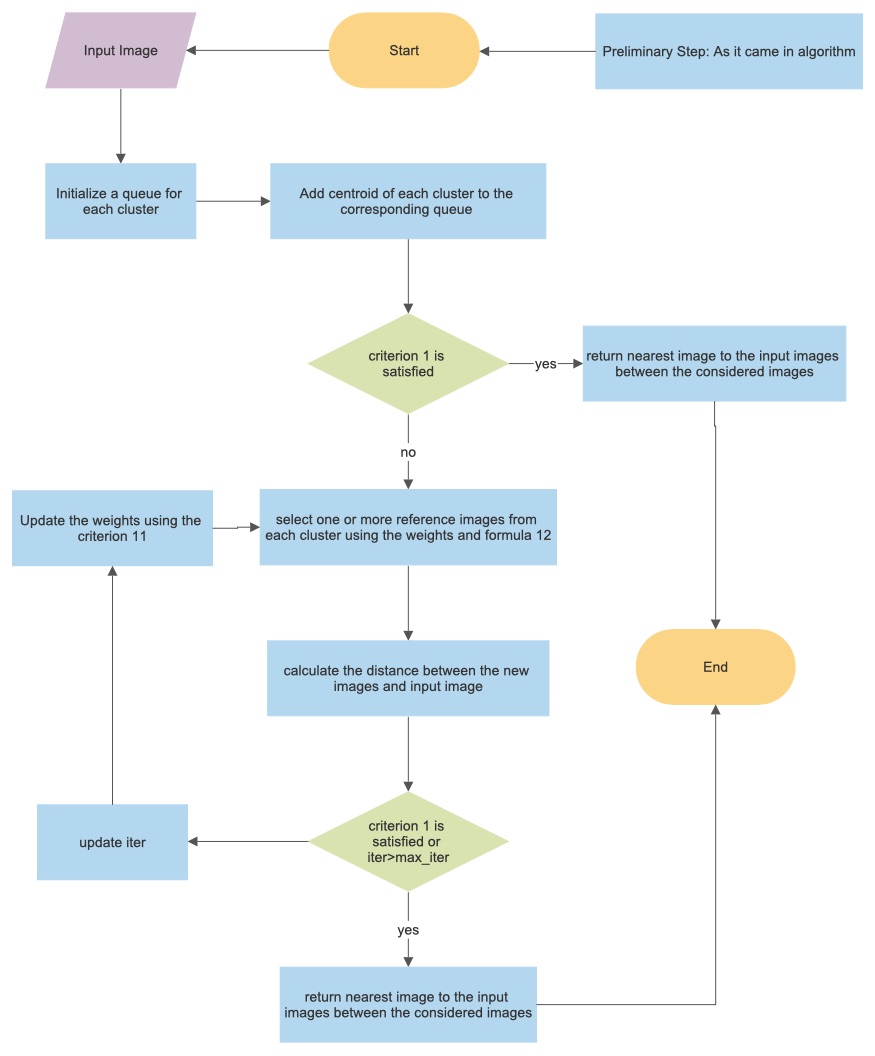}
        \captionsetup{justification=centering,margin=0cm}
        \caption{\color{black}Block diagram of proposed algorithm}
        \label{fig:1}
    \end{center}
\end{figure}

\begin{algorithm}
\color{black}
\caption{Proposed Algorithm}\label{alg:ProposedAlg}
\textbf{Data:} input image X and reference image database \{$X_r$\}\\
\textbf{Output:} The class of the nearest reference image to the input image\\
\textbf{The preliminary step:} calculating distance matrix P (P[i][j] = $\rho_{r_i,r_j}$), $\rho_0$, and clustering the reference images
\begin{algorithmic}[1]
\Procedure{Image Recognition}{input image X}
\State Initialize an array for each cluster containing their centroid
\State Initialize the array of weights
\State Calculate sum of the distances from the reference images in each queue to the input image
\State Check criterion \ref{eq:1} for all centroids
\State Update the array of weights
\State Assign iteration := 0
\While{iteration $< max\_iteration$ or criterion \ref{eq:1} is not fulfilled}
\State Update iteration
\For {each cluster in Clusters}
\State Calculate likelihood for all the images in cluster
\State Select w reference images with the bigger likelihood \Comment{w = weight[cluster]}
\State Update distances
\State Add the selected image to the queues[cluster]
\If{the selected image fulfills the criterion \ref{eq:1}}
\State {Return the selected image}
\EndIf
\EndFor
\State Update the array of weights
\EndWhile
\State Select nearest image to the input image in queues
\State Return Selected-Image
\EndProcedure
\end{algorithmic}
\end{algorithm}
\color{black}
\color{black}According \color{black}to the explanation above, \color{black}the details of the proposed method and the steps described in algorithm \color{black}\ref{alg:ProposedAlg} could be summarized as follows (the lines mentioned  in these steps are the results of the lines in the algorithm \ref{alg:ProposedAlg}): 
\begin{changemargin}{1cm}{0} 

\small \begin{steps}
\color{black}

    \item Clustering the dataset into K parts and assigning one empty queue to each cluster: $queue_i$; line: the preliminary step 
    \item Adding the centroid of each cluster to its queue; line: 5
    \item Checking criterion \ref{eq:1} for $centroid_i; i=\{1,..., k\}$ if this criterion is established, go to step 9, otherwise continue; line: 8
    \item Calculating the weights based on criterion \ref{eq:10}; line: 9
    \item Checking the number of iterations, if it exceeds the maximum number of iterations go to step 9, otherwise, continue; line: 11
    \item Selecting the instances from each cluster using \ref{eq:11} and \ref{eq:12} (the number of instances is based on the weights assigned to each cluster \ref{eq:11}); line: 13-19
    \item Checking the criterion \ref{eq:1} for each new instance; if the criterion is fulfilled return the image; line: 20-21
    \item Go to step 5
    \item Return the nearest image between the considered images; line: 24

\end{steps}
\end{changemargin}
\normalsize
Previously, we used the pseudo-code to show how the algorithm works in the algorithm \ref{alg:ProposedAlg}. \color{black}In addition, \color{black}the block diagram of the method is presented in Fig. \ref{fig:1} to provide a better view of the steps.

\section{Experimental Results}
\label{sec:3}
\begin{figure}[H]
    
     \begin{subfigure}[b]{0.4\textwidth}
         \centering
         \includegraphics[scale = 0.4]{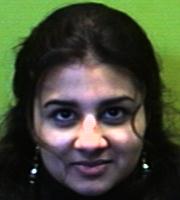}
         \caption{\color{black}An instance of dataset \cite{first_DS}}
         \label{fig:2}
     \end{subfigure}
     \hfill
     \begin{subfigure}[b]{0.4\textwidth}
         \centering
         \includegraphics[scale = 0.36]{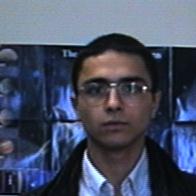}
         \caption{\color{black}An instance of dataset \cite{second_DS}}
         \label{fig:3}
     \end{subfigure}
     \captionsetup{justification=justified,margin=0cm}
     \caption[justification=justified,margin=0cm]{\color{black}instances of each datasets\ \ \ \ \ \ \ \ }
\end{figure}

\raggedbottom
In this section, we conduct a number of experiments on publicly available datasets \cite{first_DS} and \cite{second_DS}\footnote{\color{black} We use the second dataset to make a more valid conclusion} to verify the performance of our proposed method\cite{RFR}. \color{black}You \color{black} could \color{black}see an instance of datasets \cite{first_DS} and \cite{second_DS} in Figs. \ref{fig:2} and \ref{fig:3} respectively.\color{black}\\

\begin{figure}[H]
     \centering
     \begin{subfigure}[b]{0.4\textwidth}
         \centering
         \includegraphics[scale = 0.26]{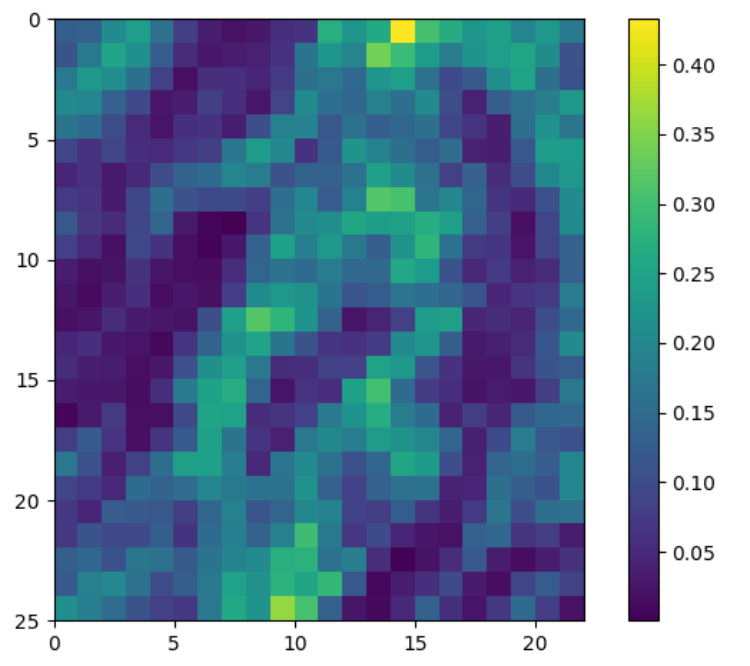}
         \caption{\color{black}Transformed form of Fig. \ref{fig:2}}
         \label{fig:18}
     \end{subfigure}
     \hfill
     \begin{subfigure}[b]{0.4\textwidth}
         \centering
         \includegraphics[scale = 0.23]{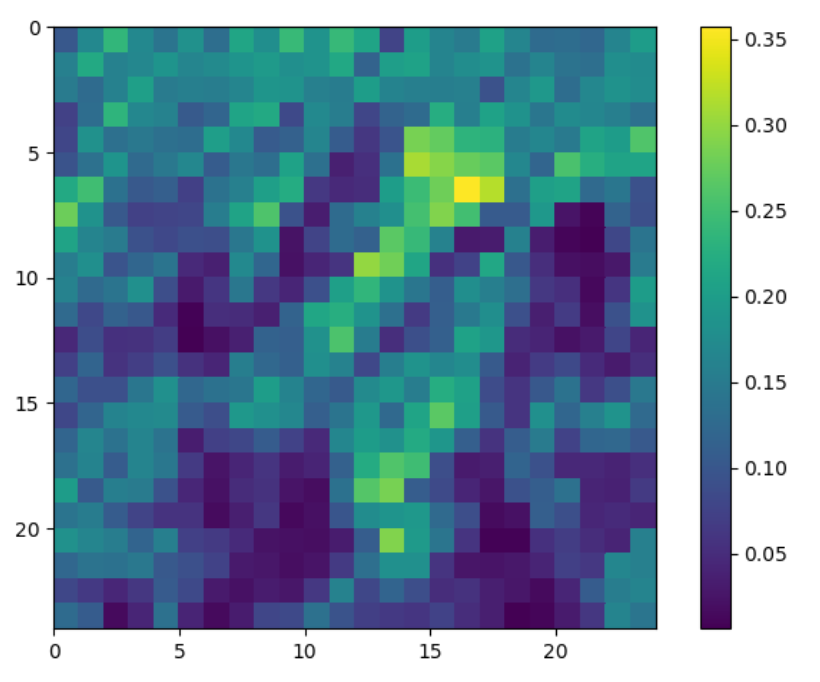}
         \caption{\color{black}Transformed form of Fig. \ref{fig:3}}
         \label{fig:19}
     \end{subfigure}
     \hfill
     \caption{\color{black}The results of carrying out HOG transformation in each instance}
\end{figure}

\color{black}This paper mainly focuses \color{black}on the face recognition problems, so the experimental study is \color{black}done on \color{black}this form of problem \cite{x}. In the experiment process, we evaluate the average recognition time and the accuracy for a different number of clusters and compare them with each other and brute force \color{black} NN \color{black}algorithm.\\ It \color{black}is worth mentioning \color{black}that the proposed algorithm is a general form of the \color{black} ML-ANN \color{black} method. In other words, when \color{black}there is \color{black} one cluster, the proposed method is the same as \color{black} ML-ANN\color{black}. \color{black}Thus\color{black}, we use the results \color{black}produced \color{black}from running the proposed algorithm with one cluster as the results of the \color{black} ML-ANN \color{black} algorithm in our comparison. \color{black} \color{black}In addition, we employ the proposed method with two and three clusters in the testing process. In the following figures, we show the ML-ANN algorithm as ``ML-ANN", which means the proposed algorithm has one cluster, and the proposed algorithm with two and three clusters is shown as ``D-ML-ANN-Cl2" and ``D-ML-ANN-Cl3". \color{black}\\
\color{black}To evaluate \color{black}the average recognition time, accuracy, and average number of distance computation, we implemented the algorithms in the python programming language and used 2 datasets to test the algorithms. In this section, the main results are obtained using the dataset \cite{first_DS} \color{black}which \color{black}contains three groups of people, \color{black}with each group including \color{black}at least 20 persons. The total number of people is 153, and there are 20 images per person, \color{black}with all images being of \color{black}size of $180 \times 200$ with RGB color space.
\\Although parallel programming makes the proposed algorithm faster, we did not use it in the experimental study.\\
As mentioned in the introduction, we used the HOG \color{black}method \color{black}\cite{HOG} to extract features; \color{black}in addition, \color{black}we used HOG descriptor class from the OpenCV library with the properties used in \cite{ML-ANN}. \color{black} The transformed form of Figs. \ref{fig:2} and \ref{fig:3} are presented in Fig. \ref{fig:18} and \ref{fig:19} respectively.\color{black}\\

\raggedbottom

We used the KMeans class from the sklearn library to do clustering. The number of clusters is based on the dataset.\\

\begin{figure}[H]
     \centering
     \begin{subfigure}[b]{0.4\textwidth}
         \centering
         \includegraphics[scale = 0.4]{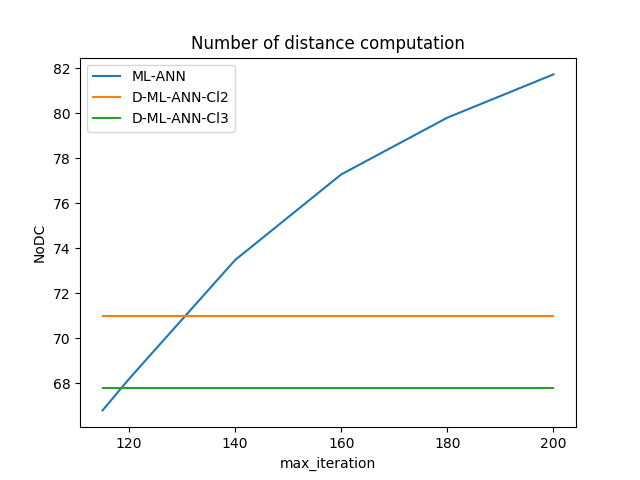}
         \caption{Threshold = 0.085, Dataset: \cite{first_DS}}
         \label{fig:8}
     \end{subfigure}
     \hfill
     \begin{subfigure}[b]{0.4\textwidth}
         \centering
         \includegraphics[scale = 0.4]{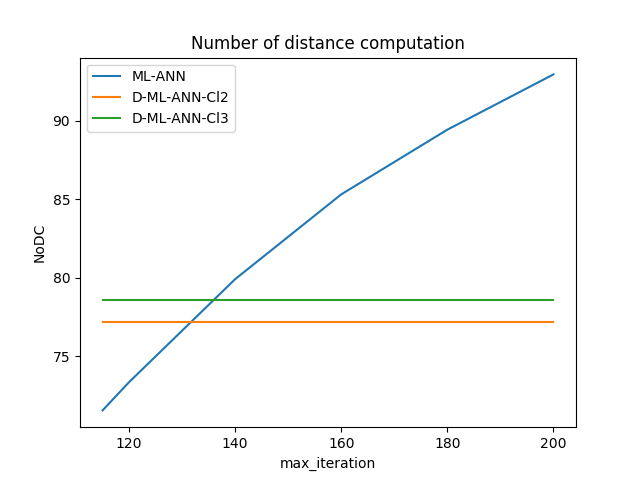}
         \caption{Threshold = 0.083, Dataset: \cite{first_DS}}
         \label{fig:9}
     \end{subfigure}
     \hfill
     \caption{\color{black}The number of distance computations per maximum iteration}
\end{figure}

\begin{figure}[H]
     \centering
     \begin{subfigure}[b]{0.4\textwidth}
         \centering
         \includegraphics[scale = 0.4]{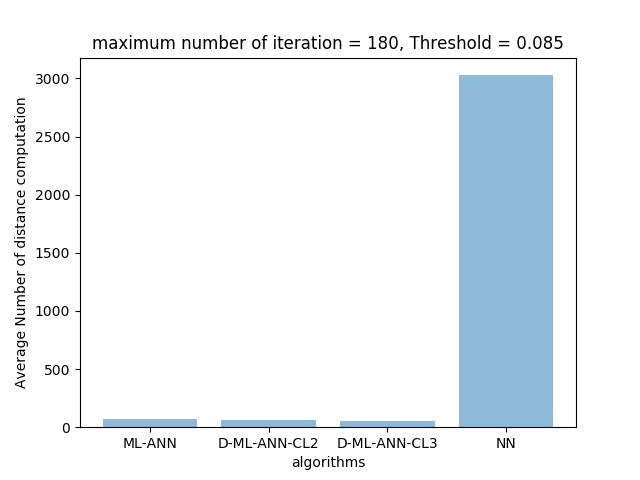}
         \caption{Threshold = 0.085, Dataset: \cite{first_DS}}
         \label{fig:4}
     \end{subfigure}
     \hfill
     \begin{subfigure}[b]{0.4\textwidth}
         \centering
         \includegraphics[scale = 0.4]{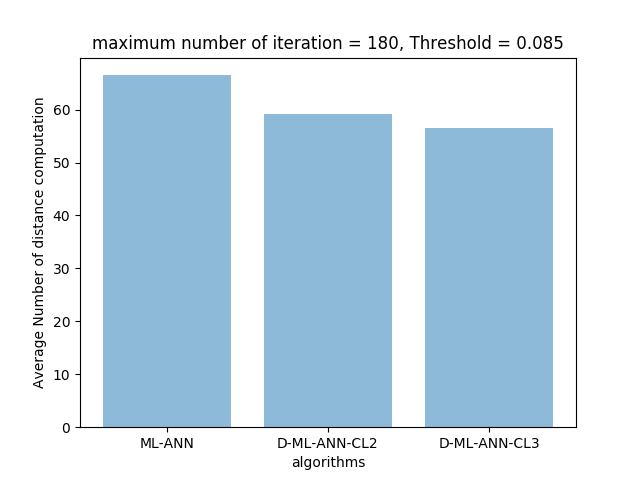}
         \captionsetup{justification=justified,margin=0.5cm,}
         \caption{Threshold = 0.085, Dataset: \cite{first_DS}}
         \label{fig:5}
     \end{subfigure}
     \hfill
     \caption{\color{black}The number of distance computation for each algorithm}
\end{figure}

We tested all algorithms \color{black}by \color{black}setting the threshold \color{black}at \color{black}0.083 and 0.085. \color{black}Besides, \color{black}we used 50 images from each dataset as input images. \color{black}By using \color{black}the term average, we mean the average value of the results generated from testing these images. \\ Since the most time-consuming part of recognition is distance computation, we \color{black}firstly \color{black}consider the average number of distance computations in each algorithm. The results \color{black}are presented in Figs. \ref{fig:8}, \ref{fig:9}, and \ref{fig:4} as follow:\color{black}\\

Both Figs. \ref{fig:8} and \ref{fig:9} \color{black}imply \color{black}that there is a great decrease in the number of distance computation\color{black}s upon using the proposed approach.\color{black}
\\Fig. \ref{fig:4} \color{black}shows \color{black}that the number of distance computations in the \color{black} NN \color{black}method is much more than that in \color{black} ANN \color{black} approaches \color{black}as expected\color{black}. \color{black}Thus\color{black}, in Fig. \ref{fig:5}, we \color{black}could \color{black}ignore the results of the \color{black} NN \color{black}method. \\
Fig. \ref{fig:5} \color{black}provides \color{black}the same information as Figs. \ref{fig:8} and \ref{fig:9}. \color{black}In addition, \color{black}it shows that the proposed approach needs less distance computations than other \color{black} ANN \color{black} approaches.\\

\begin{figure}[H]
     \centering
     \begin{subfigure}[b]{0.4\textwidth}
         \centering
         \includegraphics[scale = 0.4]{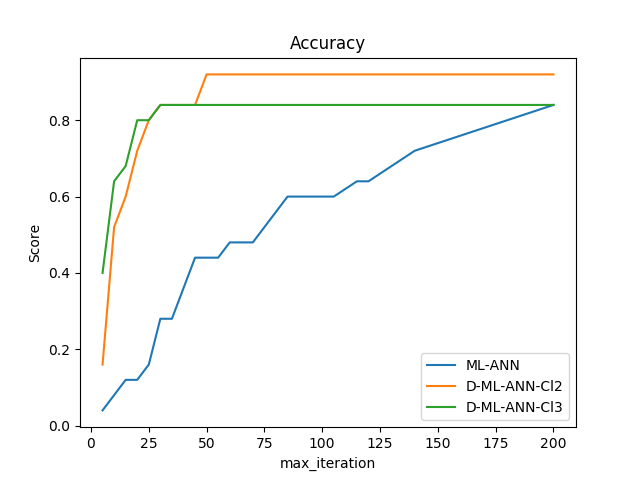}
         \captionsetup{justification=justified,margin=0.5cm}
         \caption[center]{Threshold = 0.085, Dataset: \cite{first_DS}}
         \label{fig:7}
     \end{subfigure}
    \newpage
     \begin{subfigure}[b]{0.4\textwidth}
         \centering
         \includegraphics[scale = 0.4]{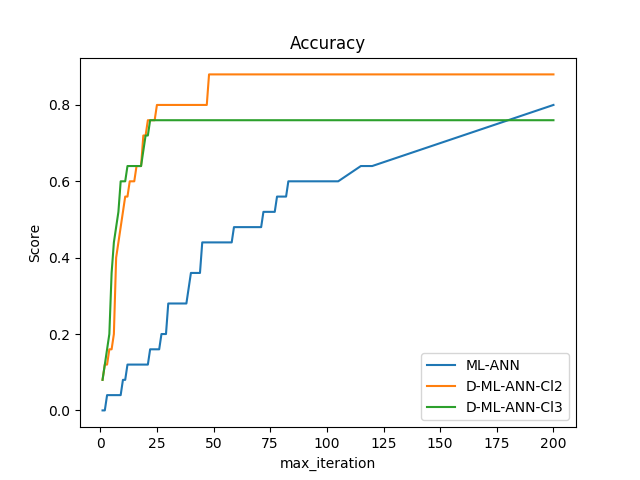}
         \captionsetup{justification=justified,margin=0.5cm}
         \caption[center]{Threshold = 0.083, Dataset: \cite{first_DS}}
         \label{fig:6}
     \end{subfigure}
     \hfill
     \caption{\color{black}Accuracy per value of the maximum iteration}
\end{figure}
\begin{figure}[H]
     \centering
     \begin{subfigure}[b]{0.4\textwidth}
         \centering
         \includegraphics[scale = 0.4]{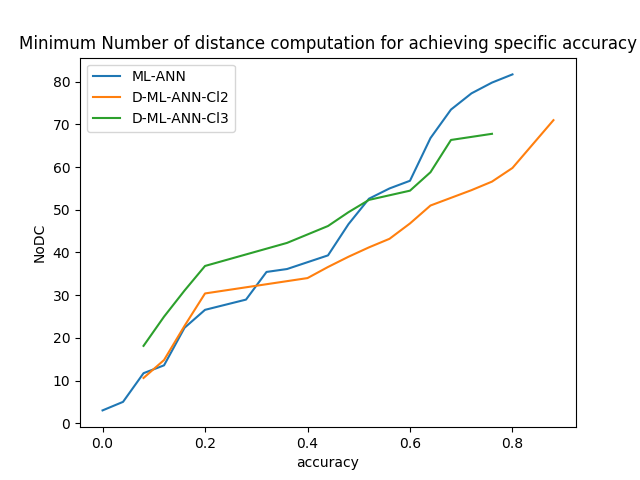}
         \captionsetup{justification=justified,margin=0.5cm}
         \caption[center]{Threshold = 0.085, Dataset: \cite{first_DS}}
         \label{fig:10}
     \end{subfigure}
     \hfill
     \begin{subfigure}[b]{0.4\textwidth}
         \centering
         \includegraphics[scale = 0.4]{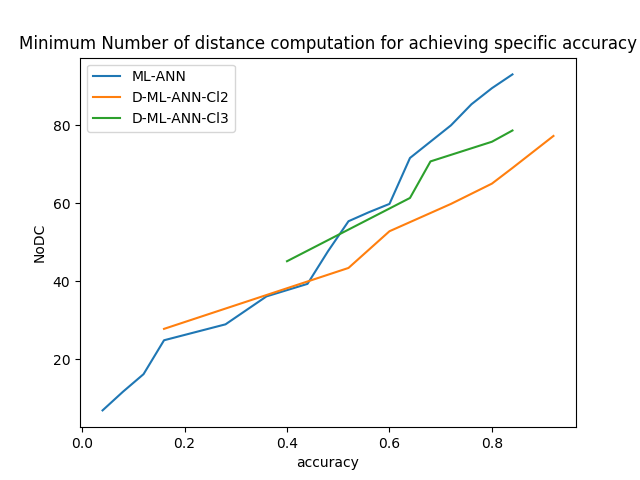}
         \captionsetup{justification=justified,margin=0.5cm}
         \caption[center]{Threshold = 0.083, Dataset: \cite{first_DS}}
         \label{fig:11}
     \end{subfigure}
     \hfill
     \caption{\color{black}The minimum value of the average number of distance computations for achieving the specific accuracy}
\end{figure}

In the following \color{black}section, \color{black}we have plotted the accuracy of each algorithm per iteration., \color{black}with Figs \ref{fig:6} and \ref{fig:7} showing the results. As \color{black}one could see in Fig. \ref{fig:6} and \ref{fig:7} there is a limitation in the accuracy, which is due to the threshold defined for the acceptable distance. Upon a decrease in this threshold, we will have more accurate outputs.\color{black}\\
As Figs. \ref{fig:6}, and \ref{fig:7} show, when maximum iterations increases, the accuracy increases as well\color{black}; however, \color{black}with the same value for the maximum number of iterations, the proposed algorithm is more accurate.\\ 
\color{black}To better understand this issue, \color{black}we evaluated the minimum\footnote{Since different values for maximum iteration could result in the same accuracy we use the term minimum to illustrate the minimum number of distance computations resulting in specific accuracy.} value of the average number of distance computations for achieving specific accuracy. The results \color{black}are presented \color{black}in Figs. \ref{fig:10} and \ref{fig:11}. \\
The results plotted in Figs. \ref{fig:10} and \ref{fig:11} demonstrate that the number of distance computations for achieving the same accuracy is fewer than other algorithms. \color{black}It must be noted \color{black}that even with the minimum number of distance computations the accuracy of proposed algorithm was higher than 0.1 with the threshold equal to 0.085 \color{black}as well as \color{black}0.5 with the threshold equal to 0.083.\\

    





\begin{figure}[H]
     \centering
     \begin{subfigure}[b]{0.4\textwidth}
         \centering
         \includegraphics[scale = 0.4]{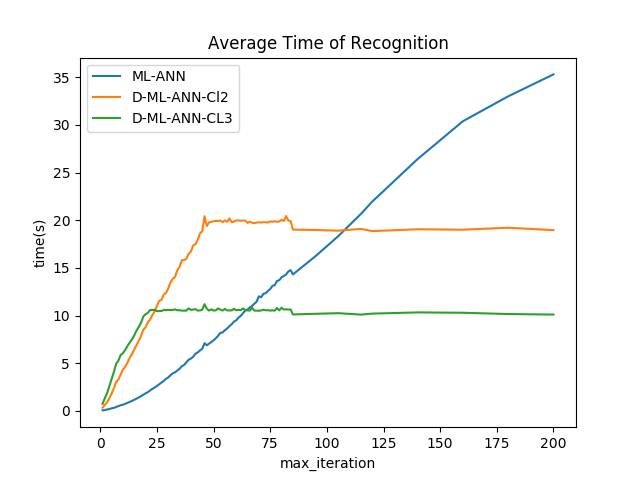}
         \captionsetup{justification=justified,margin=0.5cm}
         \caption[center]{Threshold = 0.085, Dataset: \cite{first_DS}}
         \label{fig:12}
     \end{subfigure}
     \hfill
     \begin{subfigure}[b]{0.4\textwidth}
         \centering
         \includegraphics[scale = 0.4]{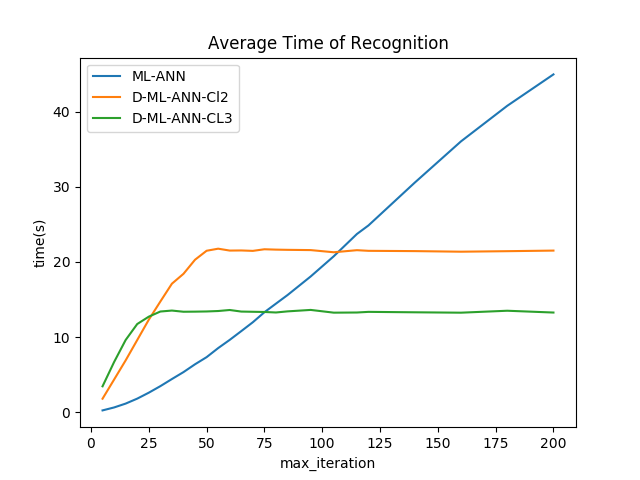}
         \captionsetup{justification=justified,margin=0.5cm}
         \caption[center]{Threshold = 0.083, Dataset: \cite{first_DS}}
         \label{fig:13}
     \end{subfigure}
     \hfill
     \caption{\color{black}The average recognition time per value for the maximum iteration}
\end{figure}

\begin{figure}[H]
    \centering
    \includegraphics[scale = 0.4]{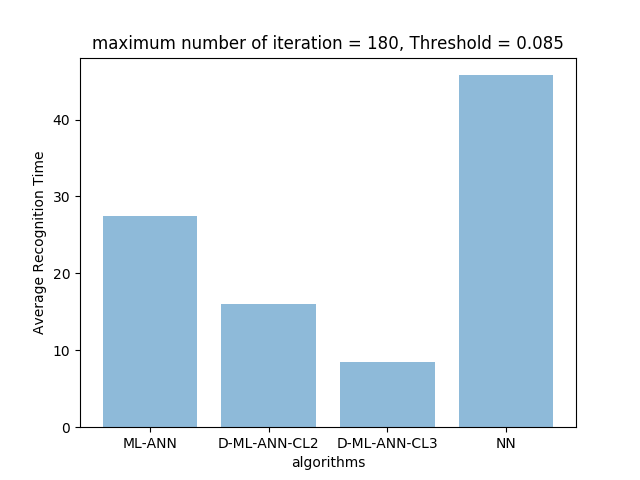}
    \captionsetup{justification=justified,margin=0.5cm}
    \caption[center]{The average recognition time per algorithm, Threshold = 0.085, Dataset: \cite{first_DS}}
    \label{fig:14}
\end{figure}

The other parameter which is so essential to our experimental study is the average recognition time. The reason for considering the average number of distance computations, was \color{black}to analyze \color{black}the recognition time in the proposed method. Therefore, we evaluated the average recognition time, \color{black}with the results shown in Figs. \color{black}\ref{fig:12}, \ref{fig:13}, and \ref{fig:14}.

Figs. \ref{fig:12}, \ref{fig:13}, and \ref{fig:14} show a \color{black}sharp decrease \color{black}in the average recognition time.

After considering the results \color{black}obtained \color{black}from testing the algorithms with the first dataset\cite{first_DS}, we also run the proposed method with different parameters on another dataset \cite{second_DS}. The number of individuals in this dataset is 152 and there is 20 images per individual. All images are of sizes $196\times196$ and the RGB color space. 
\color{black}In addition, \color{black}we evaluated the minimum value of the average number of distance computations and the recognition time for achieving specific accuracy.\footnote{\color{black}It is the same as the first study. Since, different values for the maximum iteration could result in the same accuracy, we use the term minimum to illustrate the minimum number of distance computations resulting in specific accuracy.}

\begin{figure}[H]
    \centering
    \begin{center}
        \includegraphics[scale = 0.28]{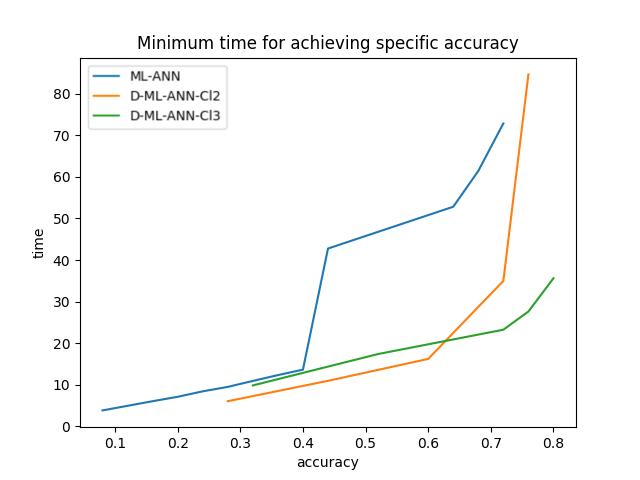}
        \captionsetup{justification=justified,margin=0.5cm}
        \caption[center]{\color{black}The minimum value of the average recognition time for achieving specific accuracy, Threshold = 0.083, Dataset: \cite{second_DS}}
        \label{fig:15}
    \end{center}
    
\end{figure}

\begin{figure}[H]
    \centering
    \includegraphics[scale = 0.28]{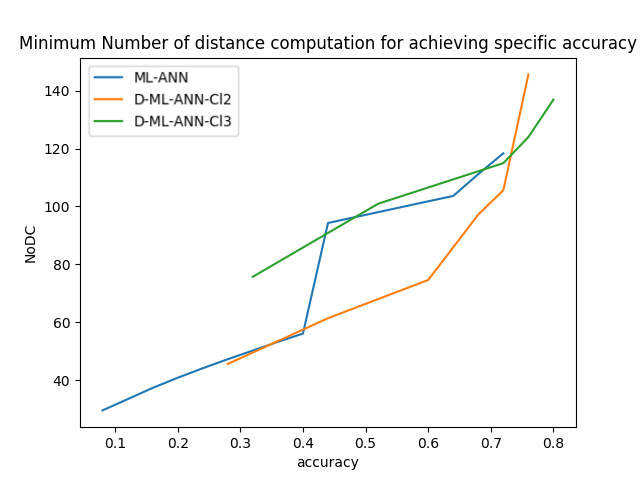}
    \captionsetup{justification=justified,margin=0.5cm}
    \caption[center]{\color{black}The minimum value of the number of distance computations for achieving the specific accuracy, Threshold = 0.083, Dataset: \cite{second_DS}}
    \label{fig:16}
\end{figure}

The comparison results were almost similar, and according to Figs. \ref{fig:15}, and \ref{fig:16}, \color{black}the performance of the proposed method was quite better in the second study as well.\color{black}\\


\section{Conclusion}
\label{sec:4}
This paper proposed an \color{black}  ANN \color{black} method, \color{black}i.e. \color{black}a generalized and distributed version of \color{black}ML-ANN \color{black}. \color{black}Accordingly\color{black}, we clustered the reference images and chose one or more instances from each cluster instead of selecting \color{black}only \color{black}one reference image from the \color{black}entire \color{black}dataset in each iteration. This process made the \color{black}selection of the \color{black}instances more conscious, \color{black}thereby enhancing \color{black}the performance of the algorithm.\\ The improvement in the algorithm \color{black}was explicitly demonstrated \color{black}in the experimental study. It is evident from the experimental results that the proposed algorithm provides better results in terms of  accuracy and the processing time.\\
As previously stated, parallel programming will make the proposed algorithm even faster. Although the \color{black}performance of the algorithm enhanced in time and accuracy, memory usage did not decrease. Hence, another way of improving the algorithm is to find a procedure for eliminating the usage of the distance matrix, thereby reducing the memory usage. \color{black}

\end{document}